\documentclass[sigplan,screen]{acmart}
\usepackage{csquotes}
\usepackage{float}
\AtBeginDocument{%
  \providecommand\BibTeX{{%
    \normalfont B\kern-0.5em{\scshape i\kern-0.25em b}\kern-0.8em\TeX}}}

\setcopyright{none}
\copyrightyear{2022}
\acmYear{2022}
\acmConference[HRI '22 PD/EUP Workshop]{PD/EUP Workshop}{March,
  2022}{Virtual}
\begin{document}

\title{End-User Puppeteering of Expressive Movements}

\author{Hongyu Wang}
\email{howiewang@cmu.edu}

\author{Nikolas Martelaro}
\email{nikmart@cmu.edu}
\affiliation{%
  \institution{Carnegie Mellon University}
  \city{Pittsburgh}
  \state{Pennsylvania}
  \country{USA}
}

% \renewcommand{\shortauthors}{Author name et al.}

\begin{abstract}
  The end-user programming of social robot behavior is usually limited by a predefined set of movements. We are proposing a puppeteering robotic interface that provides a more intuitive method of programming robot expressive movements. As the user manipulates the puppet of a robot, the actual robot replicates the movements, providing real-time visual feedback. Through this proposed interface, even with limited training, a novice user can design and program expressive movements efficiently. We present our preliminary user study results in this extended abstract.
\end{abstract}





\maketitle

%%%%%%%%%%%%%%%%%%%%%%%%%%%%%%%%%%%%%%%%%%%%%%%%%%%%%%%%%%%%%%%%%%%%%%%
%%%%%%%%%%%%%%%%%%%%%%%%%%%%%%%%%%%%%%%%%%%%%%%%%%%%%%%%%%%%%%%%%%%%%%%

\section{Introduction}
Expressive movements on robots can benefit human-robot interaction in multiple aspects including trust \cite{hamacher2016believing}, task completion \cite{reyes2015positive}, learning efficiency \cite{saerbeck2010expressive}, state indication \cite{knight2016laban}, etc. As humans, we have a keen eye for recognizing expressive movements. However, creating and programming lively robot expressions can be challenging. Often, such work is done using professional animation tools \cite{ribeiro2013nutty} or through creating code that automatically generates movements \cite{venture2019robot}\cite{bretan2015emotionally}\cite{woo2014facial}. In this project, we are investigating an intuitive interface, which features a robot and its puppet, for the end-user programming of robot expressive movement. The interface has shown effectiveness at allowing novices to create expressive movements in our ongoing user studies.

\subsection{End-User Programming of Social Robot Movements}
End-user programming on software platforms has been widely practiced. Users can define a sequence of actions on their personal computers and mobile devices to facilitate their workflow \cite{ko2011state}\cite{burnett2014future}. In the field of robotics, we have also seen previous research on end-user programming interfaces. Through technologies such as speech recognition \cite{gorostiza2011end} as well as textual \cite{buchina2016design}, visual \cite{coronado2019design}, mixed-reality \cite{gadre2019end} and tangible \cite{sefidgar2018end} programming, users with limited computer and robotics experience can command robots to complete customized tasks or movements. In fact, many interfaces of this nature are already available on the market - the Cozmo\textsuperscript{\textregistered} Code Lab \cite{cozmo} and the LEGO\textsuperscript{\textregistered} Mindstorms\textsuperscript{\textregistered} application \cite{lego_mindstorms} are recent examples.

However, for robot expressive movements, end-user programming usually involves users arranging a sequence of pre-defined robot movement elements (e.g. ``wave left arm'',  ``dance'', ``nod'', etc., as in \cite{buchina2016design}\cite{coronado2019design}). The exact designs of these elements are usually inaccessible to the end-users. Consequently, the customizability and expressiveness of the user-programmed movements are limited. These properties directly affect the perceived quality of the robot movements, which is especially important in social robots \cite{hoffman2014designing}. In addition, a social robot programming platform with limited movement customizability may potentially fail to address the cultural nuances of robot expressions \cite{trovato2012cross}\cite{ho2013interaction}.

A traditional method to program robot expressive movements is keyframing \cite{igarashi2006spatial}. Inspired by the animation technique of the same name, the designer defines the robot configuration at multiple instances when the robot completes a trajectory \cite{balit2016integrating}. The designer then either manually inputs or uses software algorithms to interpolate movements between the instances. Keyframing indeed grants a user a larger design space for expressive movements, yet the user must spend effort understanding both the robot controls and the relationship between the robot movement paths and their on-screen representations.

\subsection{Designing Robot Expressions through Puppeteering}
In search of a more intuitive, expressive, and versatile input method for end-user programming of social robot movements, we recognize the potential value of puppetry. Puppetry is an ancient art form employing various ways for humans to bring life to inanimate objects (e.g. sock puppets, rod puppets, marionettes, etc.) \cite{puppetry}. Puppets generally have direct physical control of each joint and provide immediately visible output (with the help of a mirror if necessary). For a simple puppet, even an inexperienced puppeteer can quickly create expressive movements legible to the audience.

Interestingly, the puppeteering-mirroring approach has not been extensively studied in social robot design. In one of the few relevant projects, Tennent et al. \cite{tennent2018paperino} designed a remote mirroring robotic system (PAPERINO) to conduct Wizard-of-Oz user studies on social robot behavior design. The researchers manipulated a puppet of a robot, and the robot, placed in front of the user, would copy the exact movements of the puppet. In other WoZ research, however, a variety of remote joystick-based controllers \cite{sirkin2015mechanical} \cite{bonial2017laying} or keyboards \cite{yang2015experiences} are used. These controllers lack direct correspondence \cite{martelaro2016wizard} and can make refined movement curves challenging to create \cite{mithal1996differences}.

Some earlier works, such as \cite{young2012style} and \cite{allen2012style}, proposed the concept of “Style by Demonstration (SBD)”, where a user manipulates a replica of a robot to provide stylistic movement examples for the robot to learn from. Specifically, in \cite{young2012style}, a participant moved a iRobot Roomba that was attached to a broomstick across the floor in order to demonstrate different styles of locomotion. In \cite{allen2012style}, a participant puppeteered a stuffed toy to teach its counterpart dance movements to different music. Both projects incorporated a modified reinforcement learning algorithm, originally used for animation purposes \cite{young2008puppet}. Participants of the two projects indicated their preference for the SBD approach over direct coding or non-interactive movement generation. Although these works primarily focused on robot learning, the results have shown the convenience of direct manipulation for generating robot expressive movement sequences. In the context of end-user programming, we would like to expand this interactive approach to more types of social robot behaviors, in addition to locomotion and sustained dance moves.

Therefore, we propose a robotic interface that takes advantage of puppeteering to enable end-user programming of robot expressive movements. Our goal in this project is to validate the interface prototype and acquire user feedback from different demographics. We hypothesize that even with limited training and experience in bodily movement design, a user of the interface can create a sequence of robot movements that convey emotions legible to viewers.

\section{Robot Puppeteering System}
Our robot puppeteering system is inspired by the PAPERINO system. As a proof of concept, we decided to reduce the degrees of freedom to 2 (panning and tilting) and replace the display with a wooden plate. Previous research on robot expression design has suggested that even a 1-degree-of-freedom robot can produce recognizable expressive movements \cite{bucci2017sketching}. Reducing the degrees of freedom also reduces the complexity of movement data analyses and classification.

%%%%%%%%%% Figure here
\begin{figure}[H]
    \centering
    \includegraphics[width=\linewidth]{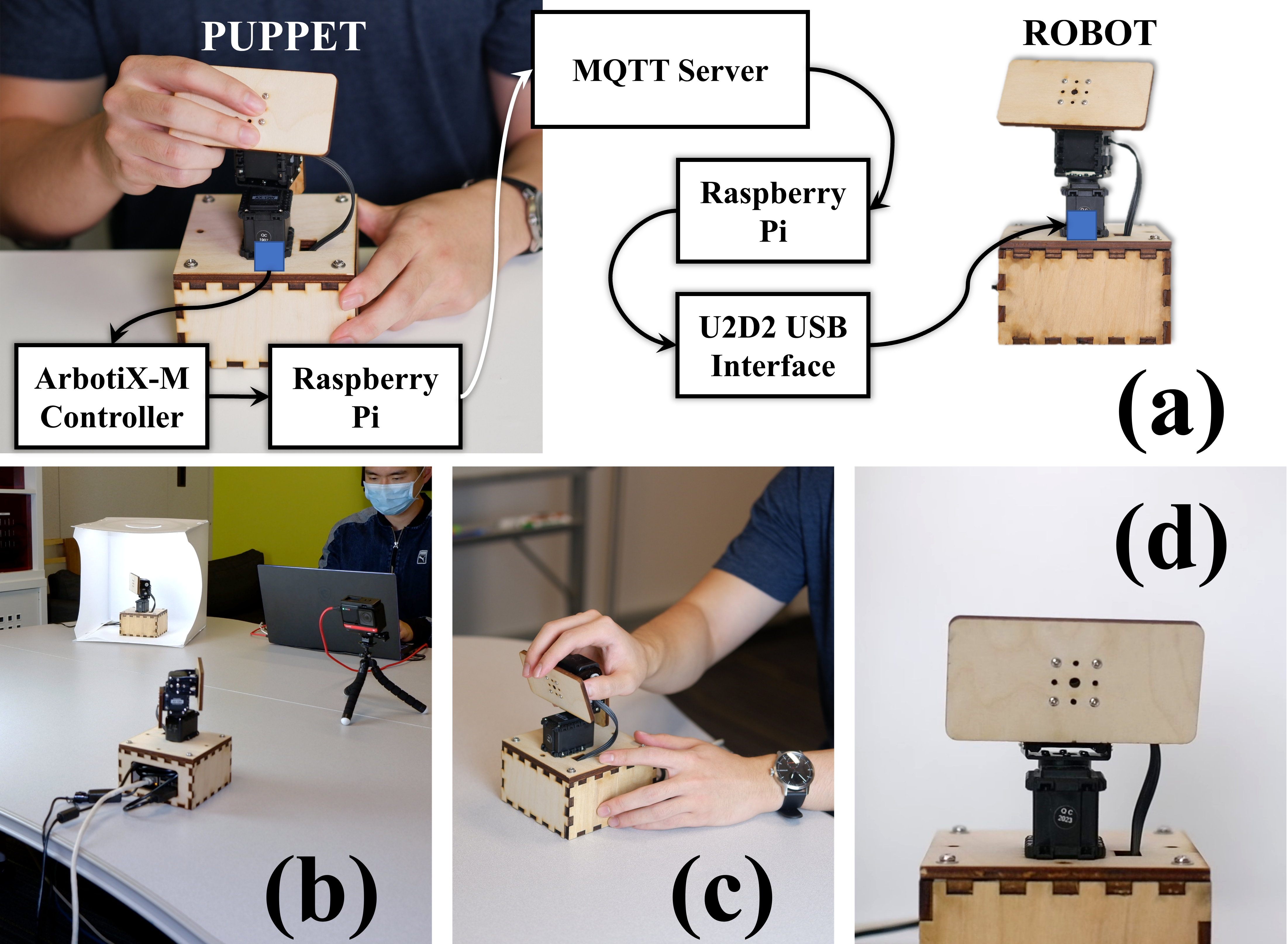}
    \caption{Robot puppeteering system setup.}
    \Description{A demonstration of the robot puppeteering system. The servo positions of the puppet are read by an ArbotiX-M controller and sent to the puppet's Raspberry Pi. The Raspberry Pi uploads the data to the MQTT server. The Raspberry Pi of the robot polls data from the server, and send the movement commands through a U2D2 USB interface to the servos of the robot. In the user study laboratory, a participants manipulates the puppet and sees the robot copying the same movement in a light box across the desk.}
    \label{fig:system}
\end{figure}

Fig. \ref{fig:system}a illustrates the 2-DOF robot and its puppet. The robot and the puppet are  connected to an MQTT server through their respective Raspberry Pis. As the user manipulates the puppet, the positions of two Dynamixel AX-12W servo motors at a fixed timestep are uploaded to the MQTT server. The robot polls the data from the server and immediately executes the motion, thereby ``mirroring'' the user inputs.

\section{Data Collection Process}
Our data collection consists of two stages: first, we obtain expressive movements generated by participants with or without experience in bodily movement (i.e. dancing, acting, etc.). In the second stage, we plan to replay the robot's movements to a separate group of participants and obtain a subjective evaluation of expression recognizability.

\subsection{Expression Generation} \label{EG}
To collect robot expressive movements created by users with different levels of experience in bodily movement design, we recruited members from our university community. Most participants had limited knowledge in robotics. Participants were invited to a user study laboratory (Fig. \ref{fig:system}b) to complete the robot expression design task\footnote{Prior to our in-person study, we validated our prototype with remote pilot studies. The participants operated the puppet at their preferred locations (home, office space, etc.) while viewing the robot movements through video conference. Despite the input lag due to network conditions, the participants were able to generate their own interpretation of the emotions. 10 out of 11 participants agreed that having the robot mirror the movement was helpful in expression design, as it provided the viewer’s perspective.}. A participant manipulates the puppet (Fig. \ref{fig:system}c) and the robot (Fig. \ref{fig:system}d), positioned under studio lighting, replicates the movements. A researcher monitored the status of the robot-puppet pair with a PC and instructed the participant to create robot expressions for six basic human emotions: anger, disgust, fear, happiness, sadness, and surprise \cite{ekman1999basic}. Each participant was compensated with a \$20 Amazon gift card.

For calibration, each expression starts with the puppet facing towards the participant’s right-hand side. The participant is asked to imagine a scenario where the robot responds to ``\textit{Hey Robot,}'' and turns towards the viewer in an emotive manner. Expressions are limited to 5 seconds. 

To start with, each participant is encouraged to "practice" their movement designs by manipulating the puppet and observing the robot movement in real time. In this way, the participant can explore the range of motion and the dynamics of the servo motors. The participant is also allowed to take some notes to aid their practice using the paper and pen placed next to them\footnote{Requested by one of the participants during the pilot study.}. When the participant finishes practicing and is ready to record one of the designs, the researcher asks the participant to execute their motion path. The movement data is recorded by the robot as it duplicates the motion of the puppet. After the first iteration of all six expressive movement designs, the participant is allowed to review their designed expressions replayed by the robot and redo unsatisfactory designs. The participant can quickly repeat the review-redesign process until they are content with the outcomes.

Once the recording session concludes, the participant completes a short survey regarding their demographic information and prior experience in expressive movement design and expression recognition. A brief interview is conducted about the participant’s thought process and user experience. 

\subsection{Expression Recognition} \label{ER}
As the second part of our data collection process, we plan to acquire subjective ratings of the collected expression designs. Another group of participants will be invited to review the expressive movements replayed by the robot using only the motion data. The replayed movements are pre-recorded in a lightbox and presented as video clips in a digital survey hosted on Qualtrics. Next to each video clip will be a list of English words describing basic emotions. After providing basic demographic information, a participant will utilize a 1-to-4 scale to evaluate the extent to which each word matches the expression in the animation. 30 different expressions designed by different expression designers will be shown.

The collected ratings will serve as an indicator of the recognizability of the designed expressive movements. We can compare the performance of expression designers with different backgrounds and levels of experience. The ratings will also be taken into account when we assign weights to these expression samples for robot learning.

\section{Preliminary Results}
We are currently expanding our collection of user-programmed expressive movements through a series of user study sessions. Some video examples of replayed movement designs can be found through this link: \url{https://bit.ly/robot-puppet-demo}.
In this section, we present some examples of user inputs and feedback.

%%%%%%%%%% Figure here
\begin{figure}[htp]
    \centering
    \includegraphics[width=\linewidth]{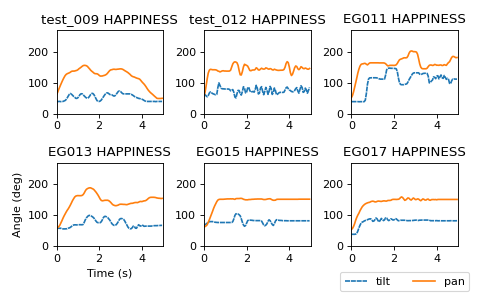}
    \caption{Examples of user-programmed expressive movements for ``Happiness''. The tilt-component (in blue dashed curves) represents the bobbing/nodding movement of the robot; the pan-component represents the horizontal shaking movement.}
    \Description{Examples of user-programmed expressive movements for ``Happiness''.}
    \label{fig:happiness}
\end{figure}

%%%%%%%%%% Figure here
\begin{figure}[htp]
    \centering
    \includegraphics[width=\linewidth]{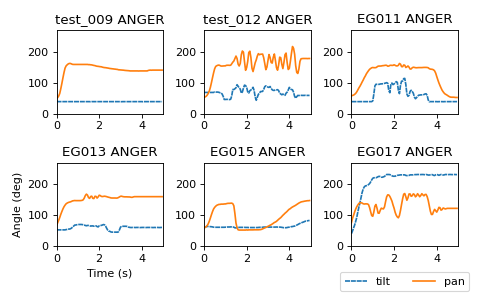}
    \caption{Examples of user-programmed expressive movements for ``anger''. }
    \Description{Examples of user-programmed expressive movements for ``anger''.}
    \label{fig:anger}
\end{figure}

\subsection{Collected Movement Data}
Fig. \ref{fig:happiness} and Fig. \ref{fig:anger} are examples of user-programmed expressive movements, whose designers had varying levels of experience in bodily movement design. From these figures, we can draw some similarities between these different interpretations of emotions. For instance, most of the programmed ``happiness'' expressions have the nodding element (i.e. undulation of the tilting component); many ``anger'' expressions involve horizontal shaking. 

For other emotions (see Appendix \ref{APPX_more_data}), we also noticed some common movement elements shared among the users, including lower tilting angles for ``sadness'' and ``fear'' (signifying low valence) and more abrupt, drastic movement curves for “surprise” (high arousal).

\subsection{User Interview}
Our interview after the Expression Generation session includes questions about the participant’s overall experience using the interface prototype and their thought process of designing expressive movements. 

Participants acknowledged their inspirations from animation characters (e.g. Wall-E and the Pixar Lamp), pets, and personal experiences and body language. One participant (EG010) mentioned that:
\begin{displayquote}
    ``It came to me that in videos, games, and… shows online, they have all sorts of robotic mechanisms. [The mechanisms] have certain actions that can express their emotions, besides the LED screens…''
\end{displayquote}

Another participant (EG016), who is an experienced ballet dancer, had an organized procedure to program the movements:
\begin{displayquote}
    ``I wrote [my designs] all out, and the first descriptor was generally faster or slower. Anger was fast. Disgust was slow… And then I decided [what] position the robot is going to end in… After I decided that, I thought about the specific journey of how to get there…''
\end{displayquote}
We can see the connection between the participant's understanding of time and space and the concept of Laban efforts \cite{knight2014expressive}\cite{laviers2012style}. Like the participant EG016, several other participants chose to take notes during the session. Excerpts of notes taken by two participants are included in Appendix \ref{APPX_user_notes}.

At least 6 out of 12 in-person participants have indicated that “disgust” is the most difficult emotion to design for, in that the emotion is “nuanced” and “complex”, involves “eyes and mouth”, and “can be misconstrued as disdain, frustration, and impatience”.

All participants agreed that having a robot mirroring the user input in real-time is helpful in designing and programming the movements, although one participant stated that depending on whom the robot is showing emotions to (e.g. the participant themselves), the mirroring may or may not be necessary. The ability to review programmed movements right after each iteration was also regarded as useful among the participants. This is in accordance with the observations in \cite{ajaykumar2020user} and the positive results from integrating movement playback into robot programming interfaces \cite{riedl2019fast}\cite{colceriu2020user}.

In addition, despite the robot prototype’s minimalist design, most in-person participants (9 out of 12 at the time of writing) of the study reported that they considered the wooden plate as the face of the robot before being introduced to the puppeteering system.

\section{Limitations and Future Work}
We will continue to investigate the similarities and differences of different users’ interpretations of human emotions applied to robots. Besides the subjective evaluation obtained from the Expression Recognition survey discussed above, we are proposing some quantitative analyses of the user-programmed movements. The analyses may include but are not limited to speed \cite{knight2014expressive}, level and frequency of peak acceleration \cite{saerbeck2010perception}\cite{lourens2010communicating}, and a more complicated analysis of speed frequency spectrum \cite{balasubramanian2011robust}. Some qualitative categorization methods based on rhythms \cite{takahashi2010remarks}\cite{koch2014rhythm} may also be incorporated for data labeling.

We also want to improve the overall user experience of this robot puppeteering system from the software perspective. A graphic user interface for recording and replay may be developed so that the users will not rely on additional instructions and controls from the researcher during the design and programming session. Additionally, we are considering incorporating visual programming into the interface: the user can intuitively program the details of individual movement "blocks" and combine the blocks to form a new sequence of movements.

We also recognize the potential for using the puppeteering concept on more complex, larger robots. 
For more complex robots with more degrees of freedom, we believe that layering recorded movements of different joints is one way to preserve the expressivity of puppeteering (similar to the overdubbing of audio tracks). For larger robots, we may utilize a scaled-down puppet suitable for hand manipulation. A previous research project has demonstrated the possibility of a remote marionette system in controlling robot posture and locomotion \cite{takubo2007wholebody}. However, we note that as the size and complexity of the robot increase, the correspondence issue may be alleviated. In this case, motion capture becomes an effective alternative input method for end-user programming. In our opinion, our puppeteering approach is currently most suitable for small-scale social robots with lower degrees of freedom.

%%%%%%%%%%%%%%%%%%%%%%%%%%%%%%%%%%%%%%%%%%%%%%%%%%%%%%%%%%%%%%%%%%%%%%%%%%%%%%%%%%%%%%%%%%%%%%%%%%%%%%%

\bibliographystyle{ACM-Reference-Format}
\bibliography{end_user_puppeteering}
\clearpage

\appendix

\section{Supplementary User Study Data}
This section presents additional materials from the Expression Generation sessions discussed in \ref{EG}.

\subsection{User-programmed robot expression examples} \label{APPX_more_data}

\begin{figure}[H]
    \includegraphics[width=78mm]{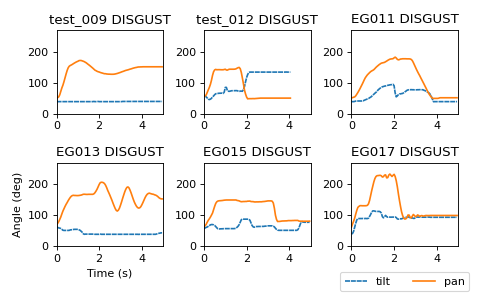}
    \caption{Examples of user-programmed expressive movements for ``disgust''.}
    \label{fig:disgust}
\end{figure}

\begin{figure}[H]
    \includegraphics[width=78mm]{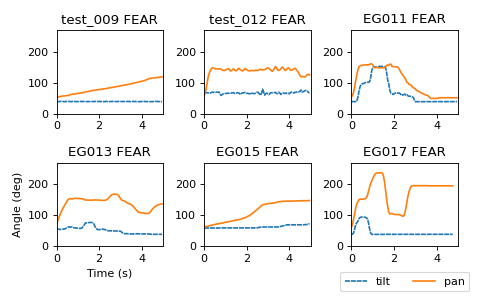}
    \caption{Examples of user-programmed expressive movements for ``fear''.}
    \label{fig:fear}
\end{figure}

\begin{figure}[H]
    \includegraphics[width=78mm]{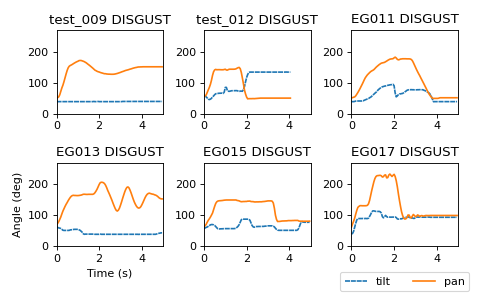}
    \caption{Examples of user-programmed expressive movements for ``sadness''.}
    \label{fig:sadness}
\end{figure}

\begin{figure}[H]
    \includegraphics[width=78mm]{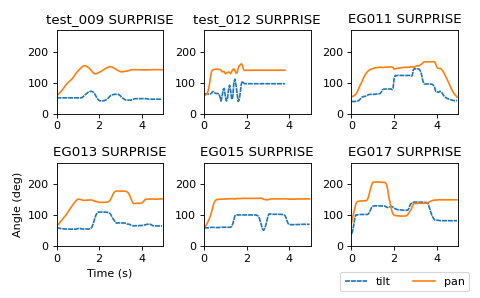}
    \caption{Examples of user-programmed expressive movements for ``surprise''.}
    \label{fig:surprise}
\end{figure}

Fig. \ref{fig:disgust} to Fig. \ref{fig:surprise} are examples of collected expressive movements, programmed by the same participants, for ``disgust'', ``fear'', ``sadness", and ``surprise", respectively.

\numberwithin{figure}{section}
%%%%%%%%%% Figure here

\subsection{Selected Notes from Participants} \label{APPX_user_notes}
\begin{figure}[H]
    \includegraphics[width=\linewidth]{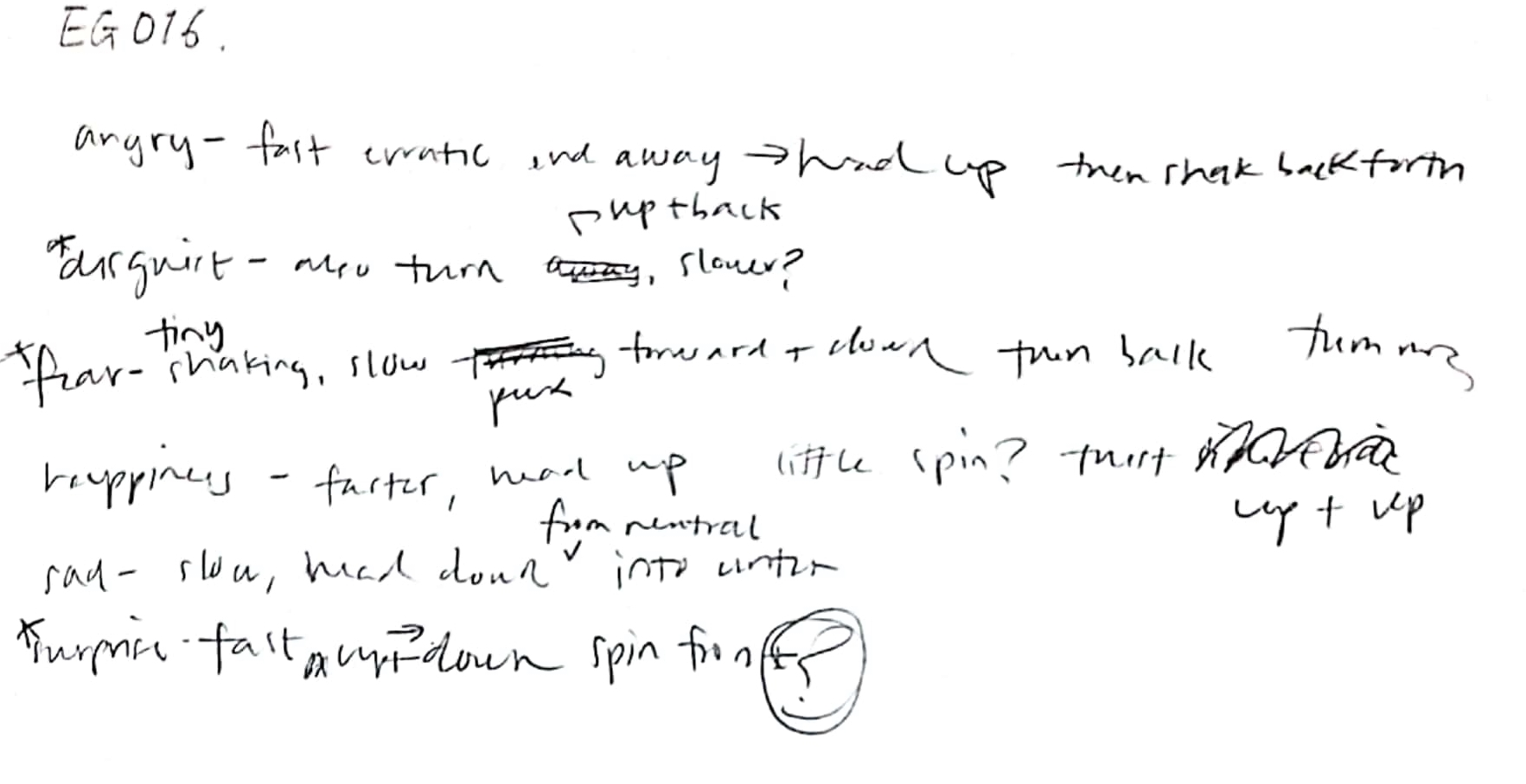}
    \caption{Textual notes taken by a participant. The participant wrote down the speed and general directions of the movements. For example, the ``happiness'' movement is ``faster,'' with ``head up'' and ``little spins(s)'' and so on. The ``sad'' movement is ``slow,'' and has ``head down from neutral into center.'' }
    \Description{Textual notes taken by a participant.}
    \label{fig:text_note}
\end{figure}

\begin{figure}[H]
    \includegraphics[width=78mm]{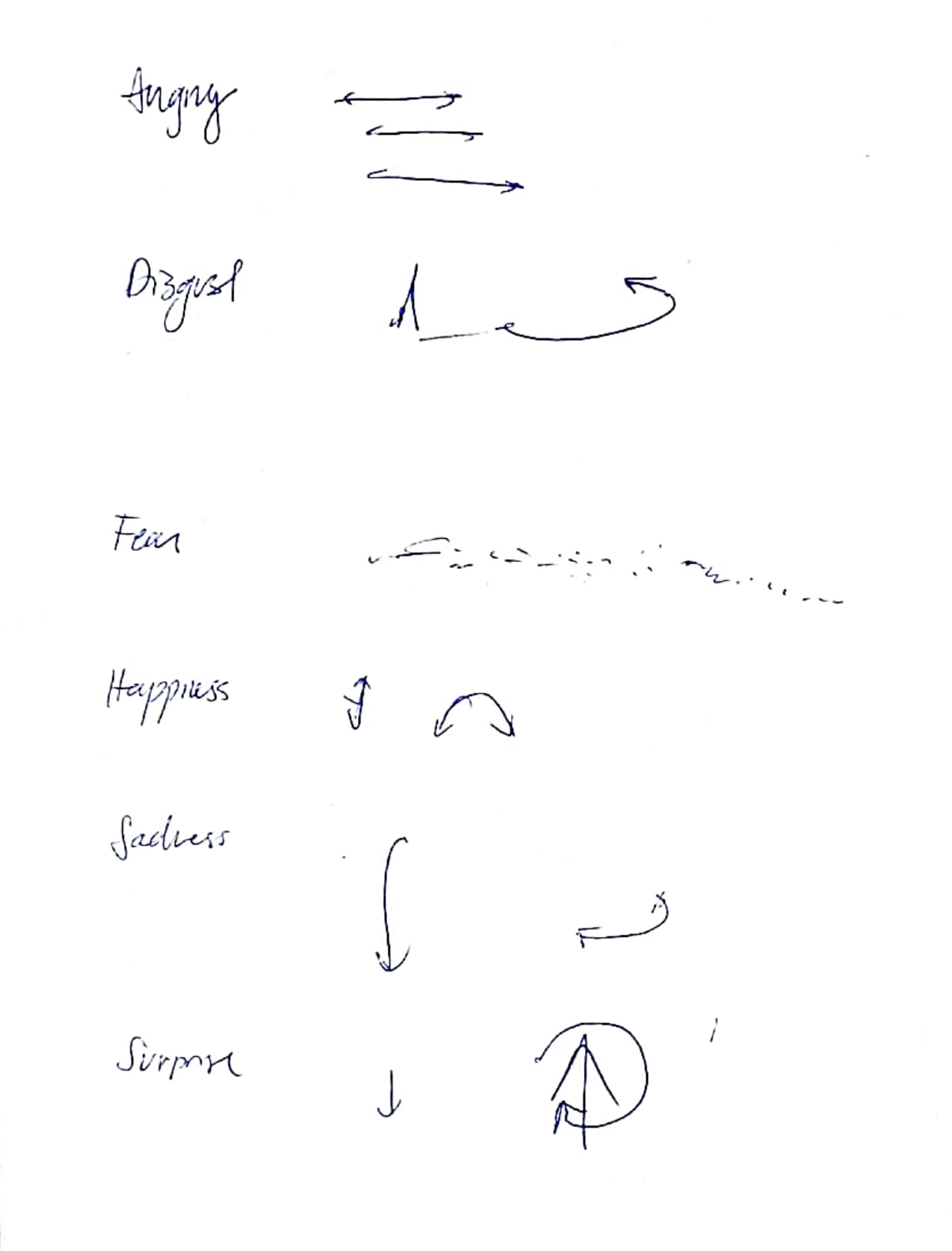}
    \caption{Graphic notes taken by a participant. The participant used arrows and lines to represent movement characters.}
    \Description{Graphic notes taken by a participant. The participant used arrows and lines to represent movement characters.}
    \label{fig:graphic_note}
\end{figure}

Fig. \ref{fig:text_note} and Fig. \ref{fig:graphic_note} are notes taken by two participants during the Expression Generation sessions discussed in \ref{EG}. A few of the participants chose to take some notes. Among these participants, some preferred to write down textual description of their movement design, while others used different lines and arrows as visual references.


\end{document}